\documentclass[11pt,a4paper]{article}
\usepackage[hyperref]{acl2020}
\usepackage{times}
\usepackage{graphicx} 
\usepackage{latexsym}
\usepackage{gb4e}\noautomath

\usepackage{microtype}

\aclfinalcopy % Uncomment this line for the final submission
\title{Overestimation of Syntactic Representation
in Neural Language Models}

\author{Jordan Kodner \\
  University of Pennsylvania \\
  Dept. of Linguistics\\
  \texttt{jkodner@sas.upenn.edu} \\\And
  Nitish Gupta \\
  University of Pennsylvania \\
  Dept. of Computer and Information Science\\
  \texttt{nitishg@seas.upenn.edu} \\}

\date{}

\newif\ifcomments
\commentstrue
\ifcomments
    \providecommand{\todo}[1]{{\protect\color{red}{[TODO: #1]}}}
    \providecommand{\jkod}[1]{{\protect\color{blue}{[JK: #1]}}}
    \providecommand{\ngup}[1]{{\protect\color{orange}{[NG: #1]}}}
\else
    \providecommand{\todo}[1]{}
    \providecommand{\jkod}[1]{}
    \providecommand{\ngup}[1]{}
\fi

\begin{document}
\maketitle
\begin{abstract}
With the advent of powerful neural language models over the last few years, research attention has increasingly focused on what aspects of language they represent that make them so successful. Several testing methodologies have been developed to probe models' syntactic representations. One popular method for determining a model's ability to induce syntactic structure trains a model on strings generated according to a template then tests the model's ability to distinguish such strings from superficially similar ones with different syntax. We illustrate a fundamental problem with this approach by reproducing positive results from a recent paper with two non-syntactic baseline language models: an n-gram model and an LSTM model trained on scrambled inputs.
\end{abstract}

\section{Introduction}

In recent years, RNN-based systems have proven excellent at a wide range of NLP tasks, sometimes achieving or even surpassing human performance on popular benchmarks. Their success stems from the complex but hard to interpret, representations that they learn from data. Given that syntax plays a critical role in human language competence, it is natural to ask whether part of what makes these models successful on language tasks is an ability to encode something akin to syntax. 

This question pertains to syntax ``in the meaningful sense,'' that is, the latent, hierarchical, largely context-free phrase structure underpinning human language as opposed to superficial or shallow issues of word order \citep{chomsky1957syntactic,marcus1984some,everaert2015structures, linzen2016assessing}. Clearly, syntactic information can be explicitly incorporated into neural systems to great effect \citep[e.g.,][]{dyer2016recurrent,swayamdipta2018syntactic}. Less certain is whether such systems induce something akin to hierarchical structure (henceforth, ``syntax'') on their own when not explicitly taught to do so.

Uncovering what an RNN is actually representing is notoriously difficult, and several methods for probing RNNs' linguistic representations have been developed to approach the problem. Most directly, one can extract finite automata \citep[e.g.,][]{weiss2017extracting} from the network or measure its state as it processes inputs to determine which neurons attend to what features \citep[e.g.,][]{shi2016does,linzen2016assessing,tenney2019bert}. Alternatively, one can present a task which only a syntactic model should be able to solve, such as grammaticality discrimination or an agreement task and then infer if a model has syntactic representations based on its behavior \citep{linzen2016assessing,ettinger2018assessing,gulordava2018colorless,warstadt2019blimp}.

In practice, simple sentences far outnumber the ones that require syntax in any natural corpus, which may obscure evaluation \citep{linzen2016assessing}. One way around this, referred to here as \textit{template-based probing}, is to either automatically generate sentences with a particular structure or extract just the relevant ones from a much larger corpus.

Templates have been used to generate data or filter examples from natural corpora for a wide range of studies, including grammaticality prediction \citep[e.g.,][]{warstadt2019blimp}, long-distance dependency resolution, and agreement prediction tasks \citep[e.g.,][]{gulordava2018colorless}.
By focusing on just relevant structures that match a given template rather than the gamut of naturally occurring sentence types, template-based probing offers a controlled setting for evaluating specific aspects of a model's representation. 

The crux of behavioral evaluation is the assertion that the chosen task effectively distinguishes between a model that forms syntactic representations and one which does not. This must be demonstrated for each task \--- if a model that does not capture syntax can pass the evaluation, then there is no conclusion to be drawn. However, this step is often omitted \citep[but not always, e.g.,][]{gulordava2018colorless,warstadt2019blimp}. Moreover, template-based generation removes the natural sparse and diverse distribution of sentence types, increasing the chance that a system might pick up on non-syntactic patterns in the data, further increasing the importance of a clear baseline. 

This problem is most clearly illustrated with an example. In the following sections, we introduce \citeauthor{prasad2019using}'s\ (\citeyear{prasad2019using}) novel psycholinguistics-inspired template-based probe of relative clause types, which was taken as evidence in support of syntactic representation in LSTMs. We then pass PvSL's test with two non-syntactic baselines: an n-gram LM which can only capture short-distance word order of concrete types (Section \ref{sec:ngram}), and an LSTM trained on scrambled inputs (Section \ref{sec:scrambled}). These baselines show that a combination of collocation and lexical representation can account for PvSL's results, which highlights a critical flaw in that experimental design. Following that, we argue that it is unlikely that LSTMs induce syntactic representations given current evidence and suggest an alternative angle for the question (Section \ref{sec:discussion}).

\section{Prasad, van Schijndel, \& Linzen 2019}
\label{sec:pvsl}
\citeauthor{prasad2019using} (PvSL; \citeyear{prasad2019using}) leverage an analogy from psycholinguistic \textit{syntactic priming} to test whether an LSTM is able to distinguish between sentences with different syntactic structures. When human subjects are \textit{primed} by receiving an example of some input, their expectation of receiving similar subsequent input will temporarily increase relative to their expectation of other inputs. This can be used to test questions about syntax because once one is primed with sentences with a specific structure, subsequent sentences with shared structure will tend to show decreased surprisal responses relative to those with different structures.

PvSL observe that this procedure may be applied to neural networks as well. Since a model's surprisal upon receiving some input decreases as it receives subsequent similar inputs, one could cumulatively ``prime'' a model by adapting it toward a certain class of input \citep{van2018neural}. As the reasoning goes, if the model can be primed for a particular syntactic structure, that implies that it is able to recognize that structure and therefore has learned a representation for it.

This paradigm is used to assess an LSTM's ability to distinguish between five superficially similar but structurally distinct sentences types: those containing an unreduced object relative clause (RC), reduced object RC, unreduced passive subject RC, unreduced passive subject RC, and active subject RC, as well as two types matched for lexical content: passive subj./obj. RC-matched coordination sentences and active subj. RC-matched coordination. To illustrate the structures, (\ref{exe:orc}-\ref{exe:src}) present an example object RC and subject RC sentence. These are distinguished syntactically by the origin of their subjects. In the first case, the subject of the sentence, `the cake,' is also the object of the relative clause (position indicated by underscore), but in the second case, the sentence subject, `the baker,' is also the subject of the relative clause. More examples can be found in PvSL \S 4.1. 

\begin{exe}
    \ex\label{exe:orc} unreduced obj. RC: \textit{The cake$_t$ }[\textit{that the baker baked \underline{\hspace{0.3cm}}}$_t$] \textit{impressed the customers.}
    \ex\label{exe:src} unreduced subj. RC: \textit{The baker$_t$ }[\textit{that \underline{\hspace{0.3cm}}$_t$ baked the cake}] \textit{impressed the customers.}
\end{exe}

As PvSL (\S 5.3) notes, if a model were able to track the position of the implicit syntactic origin, it would be able to distinguish these sentence types, so one would expect the model to exhibit a greater adaptation effect (greater decrease in surprisal) when primed and tested on the same sentence type than if primed on one type and tested on the other.

\subsection{Main Experiment}\label{sec:pvslexp}

PvSL populated templates to generate five sets of 20 adaptation and 50 test sentences for each sentence type with lexical items chosen to minimize lexical overlap between corresponding adaptation and test sets. Modifiers were optionally inserted in order to vary surface word order somewhat, and generated sentences were constrained to be \textit{felicitous}, that is, they all made plausible semantic sense.

They trained 75 LSTM language models with varying hyperparameters on five splits of the WikiText-103 corpus. Average surprisal was computed for each model for each test set, then each model was adapted to  (``primed for'') each sentence type. They were then retested on the same test sets. The difference between pre- and post-adaptation surprisal (``adaptation effect'') for each adaptation sentence type/test type pair was recorded, and adaptation effects were averaged across all models for each sentence type.

They establish a consistently and significantly stronger adaptation effect for same-type adaptation and test runs than different-type runs (PvSL \S 5.2), a stronger effect for RCs tested on models adapted for RCs rather than coordination sentences and vice-versa (PvSL \S 5.2), and for runs matched for passive voice over mismatched runs and for runs matched for reduction over mismatched runs (PvSL \S 5.4). Altogether, this is consistent with their hypothesis that the LSTM LMs are capturing abstract syntactic properties of their inputs.

Although the results are impressive, there are potential issues with their suggested interpretation. Namely, there may still be sufficient superficial word order information to achieve the effect despite the addition of optional modifiers (e.g., if unreduced object RCs often contain the bigram ``that the,'' but unreduced subject RCs never do). And, the felicity constraint means that the lexical items that appear in each sentence type should pattern together in the training data (i.e., verbs that are more likely to appear in object RCs are likely to pattern similarly in other constructions too). We test both possibilities in the following sections.

\section{N-Gram Model}
\label{sec:ngram}
We begin by training an n-gram language model (through 4-grams) with Knesser-Ney smoothing \citep{ney1994structuring} with the NLTK toolkit to determine whether it could be primed to distinguish PvSL's sentence types. An n-gram LM can only learn surface collocations and so cannot capture (hierarchical) syntax, so if it produces a significant differential adaptation effect, then the experiment is not able to discriminate between models which capture syntax from those which do not.

\begin{figure}
\begin{center}
\includegraphics[width=0.95\columnwidth]{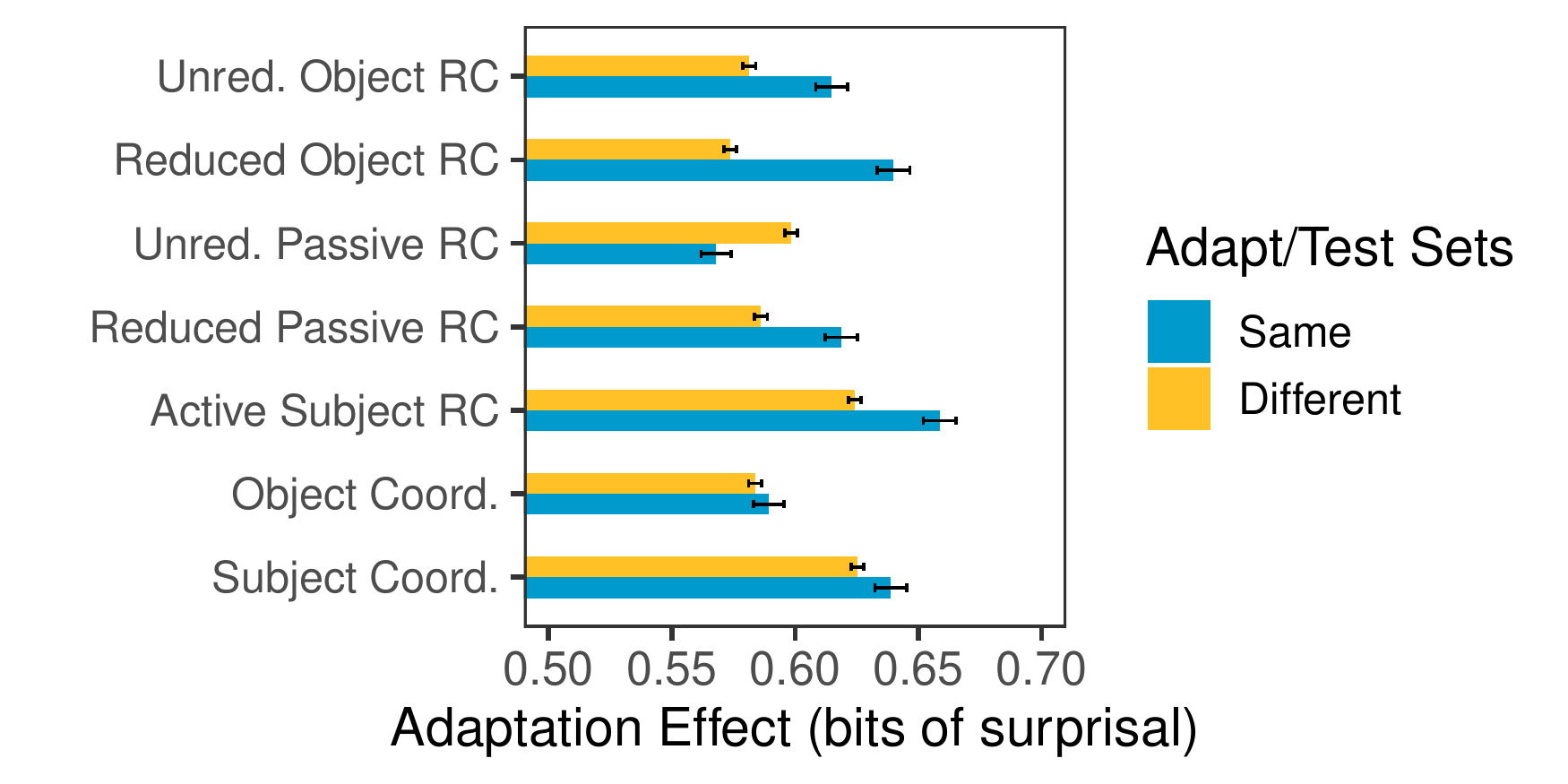}
\vspace{-0.2cm}
\caption{Average same-type vs. different-type adaptation effects for n-gram models. All differences are statistically significant except for object coordination.}\label{fig:ngramfigs1}
\end{center}
\end{figure}

Adaptation and testing were carried out with PvSL's adaptation and test sets, and LM training was modified slightly to address n-gram models' characteristics. They have no recency bias, unlike RNNs, which diminishes the impact of adaptation. As such, 20 smaller models were trained on disjoint subsets of WikiText-2 rather than the full-sized WikiText-103 subsets.

\begin{figure}
\begin{center}
\includegraphics[width=0.95\columnwidth]{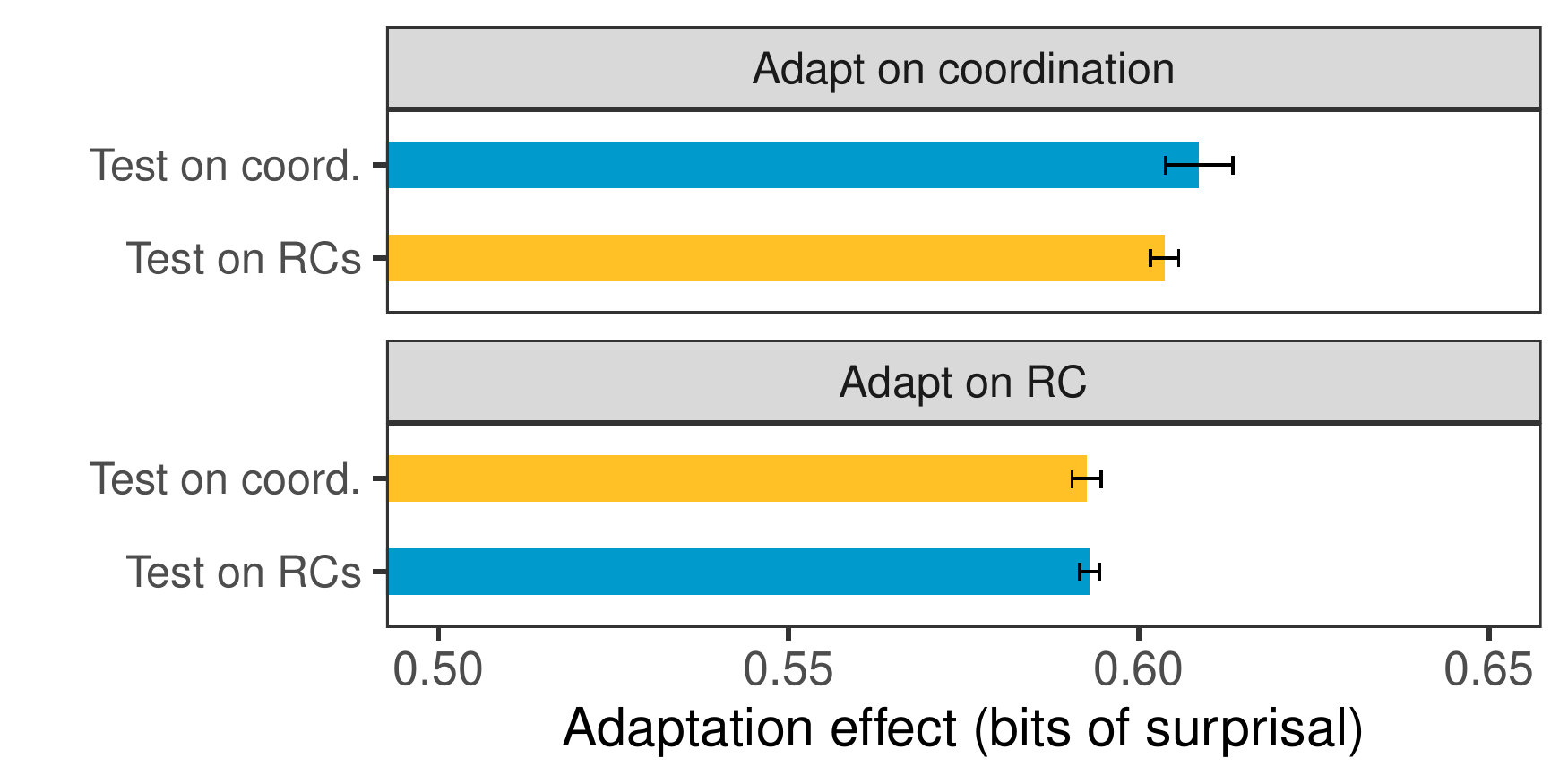}
\vspace{-0.2cm}
\caption{Average RC vs. coordination adaptation effects for n-gram models. Adapt on coord. is significant}\label{fig:ngramfigs2}
\end{center}
\end{figure}

Plotting and statistical analysis were carried out with PvSL's code\footnote{https://github.com/grushaprasad/RNN-Priming, with minor aesthetic changes to plots}. Figure \ref{fig:ngramfigs1} shows the average adaptation effect observed when the models are adapted and tested on the same sentence type or different sentence types. Importantly, the same-type adaptation effect is greater than the different-type effect for six of seven sentence types (unreduced passive RC is reversed). Although the adaptation effect is uniformly weaker than observed for PvSL's LSTM LMs, there is a statistically significant difference between the same-type and different-type effects for six of seven sentence types.

Figure \ref{fig:ngramfigs2} compares the adaptation effect over RCs compared to coordination sentences. The n-gram models show a significantly greater same-type adaptation effect for coordination but not for RCs. A small but significant increase in voice- and reduction-matched adaptation over unmatched combinations was found (matched-passive matched reduction: 0.610, matched-passive mismatched-reduction: 0.594, mismatched-passive matched-reduction: 0.575, mismatched-passive mismatched-reduction: 0.572).

\section{Scrambled-Input Model}
\label{sec:scrambled}
Next, the same \citet{van2018neural} trained LSTM LMs which PvSL employed were adapted on altered versions of their adaptation sets in which the word order of each sentence was scrambled to destroy the sentence's syntax while retaining its lexical content, then tested on the original non-scrambled test sets. Even though PvSL minimize the amount of lexical overlap in the adaptation and test sets, it may be the case that the models pick up on lexical similarities because of the felicity constraint which was imposed on them.

Scrambling was random on a sentence-by-sentence basis. Results were averaged across all the adaptation sets and models (as they were in PvSL), so the effect of any accidentally grammatical scramble was diminished. 

\begin{figure}
\begin{center}
\includegraphics[width=0.95\columnwidth]{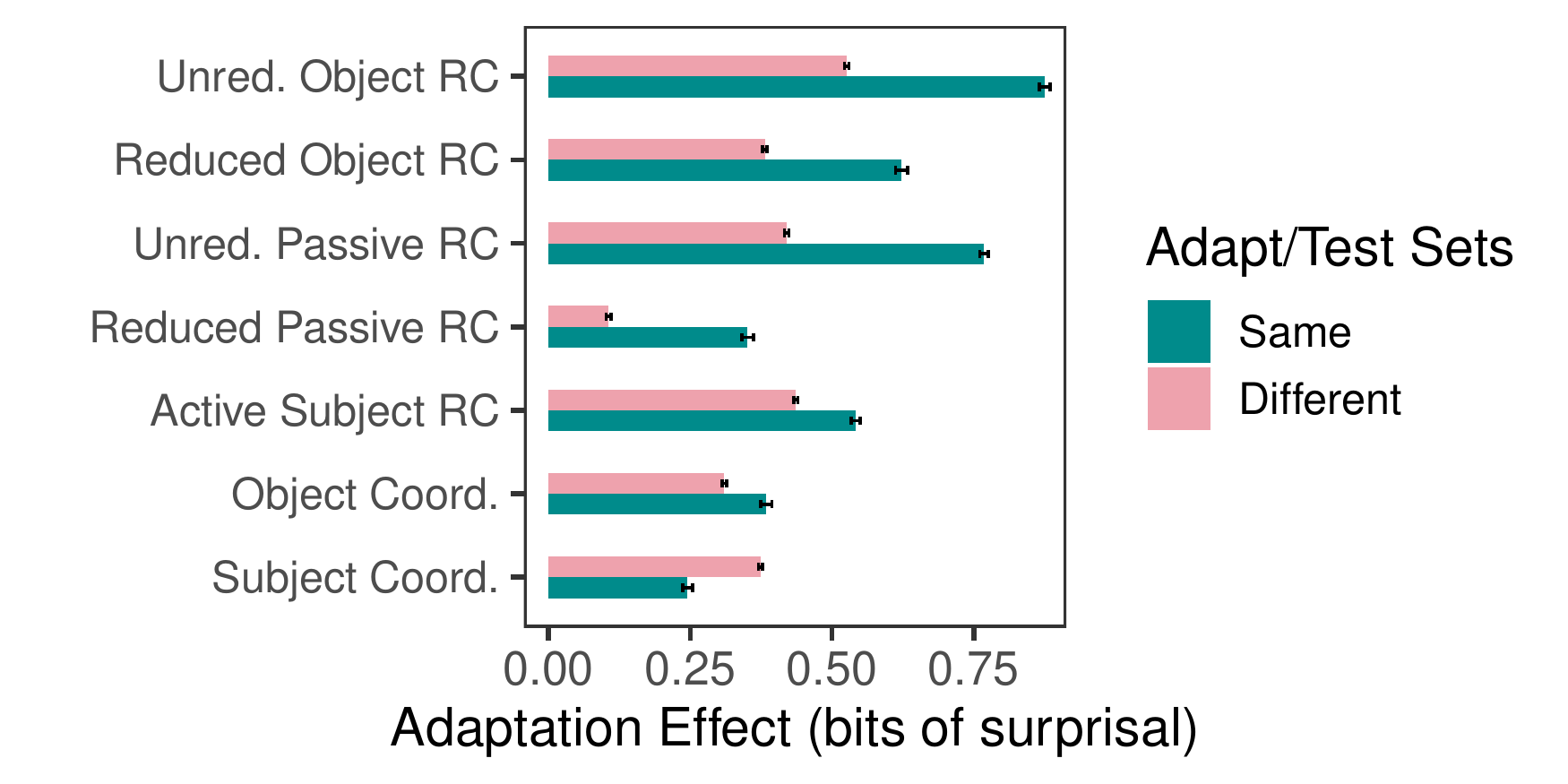}
\vspace{-0.2cm}
\caption{Average same-type vs. different-type adaptation effects for scrambled LSTM models. All differences are significant.}\label{fig:scrambledfigs1}
\end{center}
\end{figure}

Figure \ref{fig:scrambledfigs1} shows the average differential adaptation effects on these scrambled annotation runs. The same-type adaptation effect is significantly greater than different-type for six of seven sentence types (except subject coord.), and the largest relative difference is seen for unreduced passive RCs, the only type for which the n-gram models produced reverse effect. Overall, the adaptation effect is an order of magnitude larger than for the n-gram models' but still smaller than PvSL's.

\begin{figure}
\begin{center}
\includegraphics[width=0.95\columnwidth]{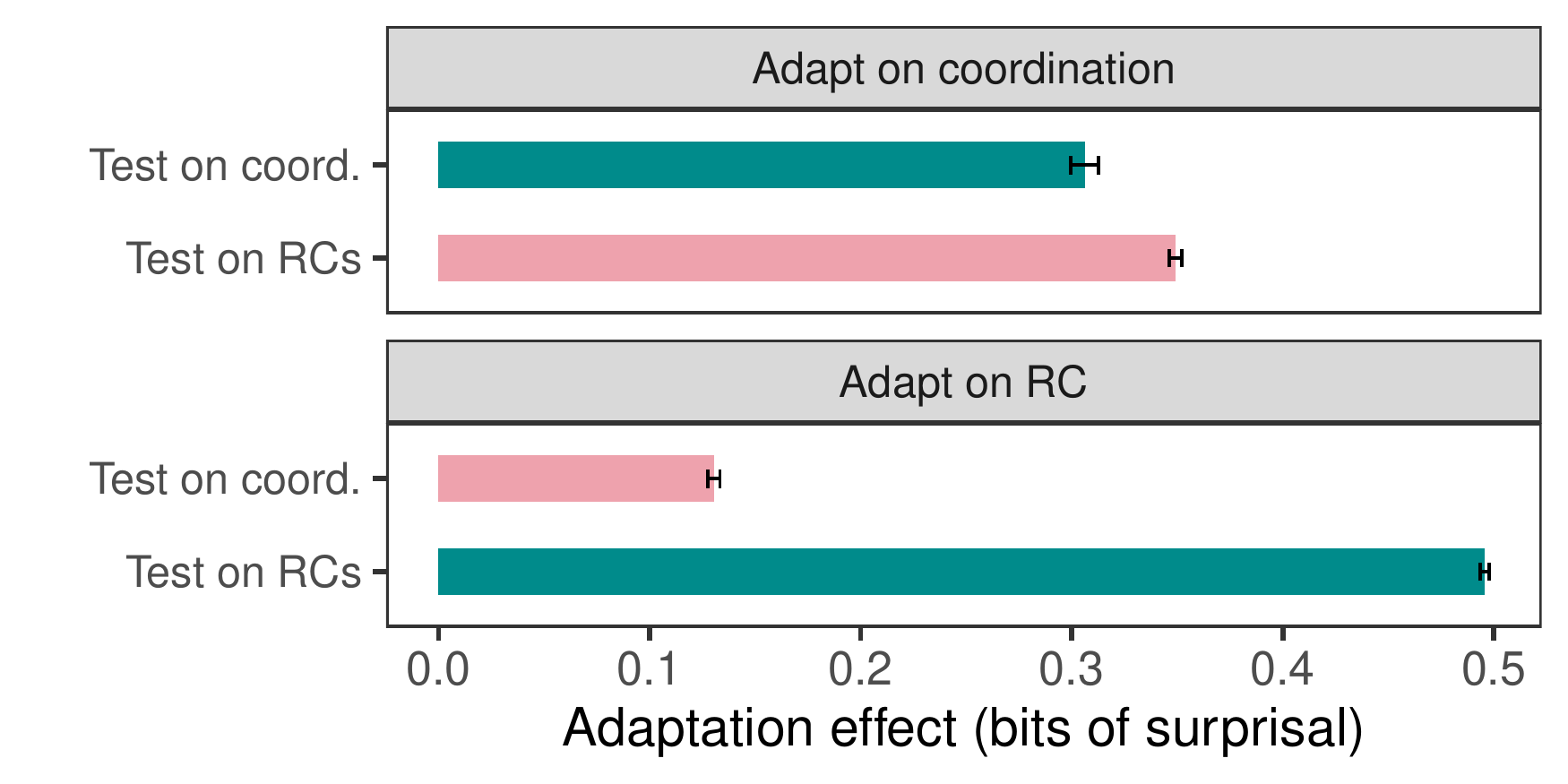}
\vspace{-0.2cm}
\caption{Average RC vs. coord. adaptation effects for scrambled LSTM models. Differences are significant.}\label{fig:scrambledfigs2}
\end{center}
\end{figure}

Figure \ref{fig:scrambledfigs2} shows differential adaptation effects for RC and coordination sentences. A backward effect is observed for sentences adapted on coordination, but a large positive effect is found for those adapted on RC sentences. This is the reverse of what was found for n-gram models. Once again, a significant positive difference was found between sentence types matched and unmatched in passives and reduction (matched-passive matched reduction: 0.65, matched-passive mismatched-reduction: 0.53, mismatched-passive matched-reduction: 0.53, mismatched-passive mismatched-reduction: 0.43).

\section{Discussion}
\label{sec:discussion}

These results call into question the \citet{van2018neural} and \citet{prasad2019using} syntactic priming paradigm's ability to distinguish models which represent syntax from those which rely on shallow phenomena by achieving a positive result with two non-syntactic baseline models. First, success in the priming paradigm is measured by whether or not adaptation reduces surprisal, but not by how much, so even though both baseline models tested here reduce surprisal by less than PvSL's models on average, they still pass the success criterion. To put it another way, PvSL report quantitative results but do not actually establish what would constitute a meaningful effect size. Even though the effect sizes of both our baseline replications were smaller, PvSL could have reported the results from our baseline models instead of their actual model and drawn the same conclusions.

Second, the fact that our surface word order n-gram model and lexical similarity-only scrambled LSTM LMs also show surprisal effects draws into question the basic claim that only a syntactic model would respond to adaptation: it is our hypothesis that the combined effect of word order and lexical similarity are what drive the LSTM models' larger effect. This is upheld, especially when it is noted that the adaptation effects of both baselines complement each other. Both alternative sources of information are well known in the community and have been tested in the past \citep{bernardy2017using,gulordava2018colorless}. This reiterates the need for proper baseline testing in computational linguistics and for informative evaluations.

This highlights a more general problem with template-based probing, namely, that the unnatural lack of sentence diversity imposed by the templates imposes unintended regularity for models to latch onto. Given the well-known observation that neural models will ``take the easy way out'' given the presence of this unintended surface information \citep{sennhauser2018evaluating, jia-liang-2017-adversarial, Naik2018StressTE}, and other work suggesting that LSTMs do not necessarily induce syntactic structure \citep{mccoy2018revisiting,gupta-lewis-2018-neural,warstadt2019blimp}, one must take successes in template-based probing studies with a grain of salt. The evaluation of non-syntactic baselines is an easy-to-implement way to combat the tendency of these behavioral probes to overestimate language models' abilities.

To improve the priming paradigm in particular, one would need to establish a success metric that discriminates between baselines and alter the experimental setup to mitigate information side channels. One possibility would be to include infelicitous ``colorless green idea'' sentences with grammatical syntax \citep[cf.][]{gulordava2018colorless}, which might decrease the lexical similarity problem. Removing the issue altogether could require enforcing completely lexically disjoint training, adaptation, and test sets, but we cannot reasonably expect a model to function when it has to generalizations to work off of, and demanding lexically distinct sets (including function words) greatly limits the set of phenomena that could be studied.

\subsection{An Alternative Approach}

As a more radical alternative, we suggest extending behavioral analysis into ``consequence-based'' analysis. From an engineering perspective, a family of models that is capable of inducing syntax is useful because it may be expected to improve performance on downstream tasks. \citet{marcus1984some} discusses in a theory-independent way which kinds of sentences a model capturing syntax should be able to parse but a ``no-explicit-syntax'' model (in the modern context, probably a baseline RNN) should not \citep[cf.][]{chomsky1957syntactic,rimell2009unbounded,nivre2010evaluation,bender2011parser,everaert2015structures}. It follows then that no-explicit- and explicit-syntax models should exhibit quantitatively different behavior on tasks that require parsing such sentences. A model that solves problems that \textit{only} one capable of inducing syntactic structure can solve may as well have induced syntactic structure from a practical standpoint. 

Consequence-based analysis would be implemented over naturalistic data rather than templates by embedding it in higher level tasks like question answering to mitigate the unnaturalness problem and demonstrate a model's practical utility. The possibility of side-channel information is already known in relation to these higher-level tasks (e.g. \cite{Poliak2018HypothesisOB, Geva2019AreWM}), and various challenge data sets have been constructed to mitigate it in different ways \citep{Levesque2011TheWS,chao2017being, dua-etal-2019-drop, Lin2019ReasoningOP,Dasigi2019QuorefAR}. Marrying these with a collection of hard sentence types \citep[e.g.,][]{marvin2018targeted,warstadt2019blimp} in something like a syntax-focused QA challenge set would provide new insights into which families of models capture the practical benefits of true hierarchical syntactic representation.

\section*{Acknowledgments}

We are particularly grateful to Marten van Schijndel for sharing the \citet{van2018neural} model checkpoints with us. We also thank Mitch Marcus, Charles Yang, and Ryan Budnick for their comments and suggestions. This work was funded by an NDSEG fellowship awarded to the first author by the ARO, in addition to funding by the ONR under Contract No. N00014-19-1-2620, and by sponsorship from the LwLL DARPA program under Contract No. FA8750-19-2-0201. (The views expressed are those of the authors and do not reflect the official policy or position of the Department of Defense or the U.S. Government.)

\bibliography{main}
\bibliographystyle{acl_natbib}

\end{document}